\documentclass{article}

\usepackage[english]{babel}
\usepackage[letterpaper,top=2cm,bottom=2cm,left=3cm,right=3cm,marginparwidth=1.75cm]{geometry}
\usepackage{amsmath, amsfonts, amssymb, amsthm}
\usepackage{graphicx}
\usepackage[colorlinks=true, allcolors=blue]{hyperref}
\usepackage{multirow}
\usepackage{array}
\usepackage{natbib}
\usepackage{hyperref}
\usepackage{stackrel}
\usepackage{here}
\usepackage{tikz-cd}
\usepackage{booktabs}
\usepackage{multirow}
\usetikzlibrary{decorations.pathreplacing}

\allowdisplaybreaks

\makeatletter
\def\paragraph{\@startsection{paragraph}{4}{\z@}{1.5ex plus 0.5ex minus .2ex}{-1em}{\normalsize\bf}}
\makeatother

\theoremstyle{plain}
\newtheorem{theorem}{Theorem}
\newtheorem{definition}{Definition}
\newtheorem{lemma}{Lemma}

\title{Is Smoothness the Key to Robustness? A Comparison of Attention and Convolution Models Using a Novel Metric}
\author{Baiyuan Chen\thanks{The University of Tokyo. Email: \href{mailto:chenbaiyuan75@g.ecc.u-tokyo.ac.jp}{\texttt{chenbaiyuan75@g.ecc.u-tokyo.ac.jp}}}}

\begin{document}
\maketitle

\begin{abstract}
Robustness is a critical aspect of machine learning models. Existing robustness evaluation approaches often lack theoretical generality or rely heavily on empirical assessments, limiting insights into the structural factors contributing to robustness. Moreover, theoretical robustness analysis is not applicable for direct comparisons between models. To address these challenges, we propose \textit{TopoLip}, a metric based on layer-wise analysis that bridges topological data analysis and Lipschitz continuity for robustness evaluation. TopoLip provides a unified framework for both theoretical and empirical robustness comparisons across different architectures or configurations, and it reveals how model parameters influence the robustness of models. Using TopoLip, we demonstrate that attention-based models typically exhibit smoother transformations and greater robustness compared to convolution-based models, as validated through theoretical analysis and adversarial tasks. Our findings establish a connection between architectural design, robustness, and topological properties.
\end{abstract}

\section{Introduction}

Robustness is a fundamental aspect of machine learning models~\citep{bai2021transformers,wang2022can}. Building a robust model has various advantages, which include maintaining high performance under various input corruptions, being resilient to adversarial attack, and generalizing well to out-of-distribution data~\citep{buzhinsky2023metrics, szegedy2013intriguing, boopathy2019cnn}. When evaluating the robustness of models, performance analysis or intrinsic robustness analysis are mainly used~\citep{weng2018evaluating, wang2016theoretical, hein2017formal}. Performance analysis assesses how well a model maintains metrics like accuracy and prediction entropy under adversarial attacks, noise, or other perturbations~\citep{carlini2017towards, rathnakumar2024bayesian}. While this approach provides straightforward empirical evaluations, it often lacks insights into the theoretical and structural properties contributing to robustness. In contrast, intrinsic robustness analysis examines properties such as the Lipschitz continuity of models, probability distributions, or intrinsic dimensions~\citep{szegedy2013intriguing, buzhinsky2023metrics, tulchinskii2024intrinsic}. However, such techniques are either difficult to generalize (to multi-layer or other complex settings), or the given bounds are too loose. Furthermore, such methods do not allow direct theoretical robustness comparison across different architectures or configurations. Lacking theoretical foundations and relying largely on experiments for robustness analysis may lead to inconsistent conclusions and limited insights into the underlying mechanisms contributing to robustness.

To fill this gap, we propose a method for robustness comparison. By comparing the robustness of different architectures or configurations, we can gain insights into choosing or developing robust models without deriving exact robustness based on some metric. Based on the comparison method, we propose \textit{TopoLip}, a metric based on layer-wise analysis that enables robustness comparison in both theory and experiments. Additionally, TopoLip provides insights into how model parameters influence robustness. 

To introduce and validate the proposed comparison method, we use Transformer and ResNet $-$ two dominant classifiers in vision tasks using distinct approaches $-$ for comparison. The Transformer architecture, introduced by \citet{vaswani2017attention}, has become highly popular and has made significant impacts across various fields. In contrast, ResNet, introduced by \citet{he2016deep}, is built using convolutional layers with residual connections. As noted by \citet{bai2021transformers}, Transformers are more robust than CNNs when handling out-of-distribution data. We further validate this by theory and experiments.

This paper is organized as follows: Section~\ref{sec_preliminaries} presents preliminary concepts; Section~\ref{sec_topolip} introduces the TopoLip robustness metric; Section~\ref{sec_wass_lip} demonstrates that attention-based models are smoother and thus more robust than convolution-based models; and Section~\ref{sec_exp} provides experimental results to validate our theoretical findings. Our main contributions are:
\begin{itemize} 
    \item We propose a metric for evaluating the robustness of models. This metric allows for both theoretical and empirical comparisons of robustness across models and provides insights into how model parameters influence robustness.
    \item We establish a relationship between the Lipschitz continuity of persistence diagrams and probability distributions, linking topological data analysis to robustness evaluation.
    \item We analyze the mean-field regime of attention and convolution layers, comparing their Lipschitz conditions. Our findings reveal that attention layers are inherently more robust to variations in input data distributions compared to convolution layers.
    \item We extend the analysis to Vision Transformers (ViTs) and ResNets, demonstrating a consistent relationship between architectural design and robustness.
    \item Finally, we validate our theoretical insights through adversarial tasks. Experimental results confirm that attention-based models generally exhibit greater robustness than convolution-based models in handling corrupted data.
\end{itemize}

\subsection{Related work}
\paragraph{Robustness metric.}
\citet{weng2018evaluating}~converts robustness analysis into a local Lipschitz constant estimation problem to derive theoretical robustness. However, the method is algorithm-based, and the process relies on computing some metric that serves as a theoretical performance for the final robustness derivation. Therefore, it can not derive a robustness score without running the algorithm in a concrete setting, making it impossible to get insights into the model robustness completely by theory. Similarly, \citet{weng2018evaluating} developed a robustness metric that is attack-independent and can be used with any neural network classifier. However, this approach is not well-suited for the theoretical analysis of individual models. \citet{buzhinsky2023metrics} proposes a metric to measure the robustness of a classifier. This metric is based on probabilistic reasoning within the latent spaces of generative models, which makes it challenging to apply to specific model settings. \citet{hein2017formal}~derives a closed-form Lipschitz bound for evaluating the robustness of a multi-layer perception (MLP) with a single hidden layer. Nevertheless, a closed-form bound is hard to derive
for a neural network with more than one hidden layer, not to mention increasing the complexity of the architecture such as the transformer. \citet{wang2016theoretical}~uses topology to study robustness. However, no robustness bounds or estimates were provided for neural networks, and thus no comparison can be made between architectures or model configurations. 

Here, we propose a method that can not only derive a Lipschitz bound for robustness completely by theory, but the bounds of different architectures or configurations can be compared with each other by replacing the values of parameters as well. Moreover, the comparison is not only robust in theory but also in experiments.

\paragraph{Topological Data Analysis.} This work is partly built upon Topological Data Analysis (TDA), which focuses on measuring the topological structures within data. The Wasserstein distance is extensively used in TDA to quantify differences between the topological structures of distributions \citep{cohen2005stability}. Although persistence diagrams (discussed in Appendix~\ref{appendix_ph}) are not equivalent to probability spaces, they possess properties that allow for the definition of probability measures \citep{mileyko2011probability}. In our study, we further explore the relationship between persistence diagrams and probability spaces, particularly in terms of their Lipschitz continuity.

\section{Preliminaries}
\label{sec_preliminaries}

\subsection{Problem setup}
Suppose the input is a 2D image with resolution $(H, W)$ and $C$ channels. In Vision Transformers (ViT), the image is reshaped into a sequence of flattened patches $\textbf{p}\in\mathbb{R}^{N\times (P^2\cdot C)}$, where $(P, P)$ is the resolution of the patches and $N=HW/P^2$ is the number of patches~\citep{dosovitskiy2020image}. This input is then mapped by an embedding matrix $\textbf{E}\in\mathbb{R}^{(P^2\cdot C)\times d}$, where $d$ is the embedding dimension. The mapping yields a matrix of size $\mathbb{R}^{N\times d}$, which can be interpreted as a sequence of $N$ input vectors $\{x_i\}_{i=1}^N\subset \mathbb{R}^d$. These vectors are often expressed as an input matrix $X = [x_1, ..., x_N]\in\mathbb{R}^{d\times N}$.

For a convolutional layer in residual networks (ResNet), let $\mathbf{y}(\alpha)\in\mathbb{R}^{C}$ represent the input at position $\alpha$. By utilizing a $(2k+1)\times(2k+1)$ filter, the response of a convolutional layer at position $\alpha$ can be written as $\overline{\mathbf{y}}(\alpha) = \sum_{\beta\in ker}W_{::,\beta}\phi(\mathbf{y}(\alpha+\beta))+b$, where $W\in\mathbb{R}^{C\times C\times (2k+1)^2}$ is a weight matrix representing $C$ filters (where we set the $\#(filter)=\#(channel)$), $\phi$ denotes the activation function, and $b\in\mathbb{R}^C$ is a bias term. Since there are $H\times W$ positions at the input image, each corresponding to one response, the input image can be regarded as a $C\times N'$ sequence where $N'=HW$. More details of the convolutional layer setting will be discussed later.

Previous works have restricted the input sequence of the attention layer $X=[x_1,...,x_N]\in B_R^N$ where $B_R\subset \mathbb{R}^d$ is the closed ball centered at 0 and of radius $R$~\citep{castin2024smooth, geshkovski2024emergence}. We apply this restriction and assume each dimension of $x_i\,\,(i\in[N])$ is drawn i.i.d. from $N(0,\sigma^2)$. Specifically, by applying Chebyshev's inequality that with high probability $1-d/t^2$, we have $\|x_i\|\leq t\sigma$. For the convolution layer, we assume the input $Y=[y_1,..., y_{N'}]\in B_R^{\prime N'}$ where $B_R^\prime\subset \mathbb{R}^{C}$. Since we set $C$ infinitely large to introduce the mean-field regime of convolution, we instead bound each element: with a high probability $1-1/t^2$, we have $|y_{ij}|\leq t\sigma$.

\subsection{Discrete frameworks}

We define the discrete frameworks of attention and convolution same as the settings in the previous research \citep{he2015delving, chi2023latent}. 

\begin{definition}[Attention layer]
\label{def_attn}
Given an input sequence \(X \in \mathbb{R}^{d \times N}\), consider a single-head attention layer with parameters \( \{Q_m, K_m, V_m\}_{m\in[M]} \subset \mathbb{R}^{d \times d}\). The output of the single-head attention layer is denoted as \(\overline{X} = \textnormal{Attn}_m(X) = [\overline{x}_1, \dots, \overline{x}_{N}] \in \mathbb{R}^{d \times N}\), where each \(\overline{x}_i\) for \(i \in [N]\) is given by

\[
\overline{x}_i = \sum_{j=1}^{N} \text{softmax}\left(\frac{x_i^\top Q_{m}^\top K_{m}x_j}{\sqrt{d/M}}\right) V_mx_j = \sum_{j=1}^{N}\frac{\exp \left( x_i^\top Q_{m}^\top K_{m}x_j/\sqrt{d/M} \right)}{\sum\limits_{k=1}^{N} \exp \left( x_i^\top Q_{m}^\top K_{m}x_k/\sqrt{d/M} \right)}V_mx_j.
\]

A multi-head attention extends this concept by allowing the model to attend to information from different representation sub-spaces jointly. A $M$-head attention layer is defined as $\textnormal{MHAttn}(x_i,X):=\textbf{o}_i,$ where

\[
\textbf{o}_i = W^O(\oplus_{m=1}^M \text{head}_m )
\]
\[
\text{head}_m = [\textnormal{Attn}_m(X)]_{:i} = [\textnormal{Attn}(X;\{Q_m, K_m, V_m\})]_{:i},
\]
with \( W^O \in \mathbb{R}^{d \times Md} \) being learned projection matrices, and \( [A]_{:i} \) denotes the \( i \)-th column of matrix \( A \).
\end{definition}

Next, we define the Transformer with PreLayer Normalization (Pre-LN), which is used in various systems~\citep{fan2019deep, katharopoulos2020transformers, xiong2019layer}. For a given input vector \( x_i \in \mathbb{R}^d \), layer normalization transforms it as
\(\mathrm{LN}(x_i) = (x_i - \mu_i)/\sigma_i \odot \gamma + \beta\), 
where $\mu_i = 1/d \sum_{j=1}^{d} x_{i,j}$, $\sigma_i = \sqrt{\sum_{j=1}^{d} (x_{i,j} - \mu_i)^2/d}$, \( \gamma \in \mathbb{R}^d \) and \( \beta \in \mathbb{R}^d \) are learned scaling and shifting parameters, and $\odot$ denotes element-wise multiplication. An MLP layer with hidden dimension $d'$ is defined as
\(\mathrm{MLP}(x_i) =  W_2\phi(W_1 x_i+b_1)+b_2\)
where \(W_1 \in \mathbb{R}^{d'\times d},  W_2\in\mathbb{R}^{d \times d'}, b_1\in\mathbb{R}^{d'}, b_2\in\mathbb{R}^{d}\), and $\phi$ denotes the ReLU function. The Pre-LN Transformer is then expressed as:
\[
    \mathrm{TF}(X) = \mathrm{MLP}\circ \mathrm{LN} \Bigl(X + \mathrm{MHAttn}\circ\mathrm{LN}(X)\Bigr) + \mathrm{MHAttn}\circ\mathrm{LN}(X) + X.
\]

\begin{definition}[Convolutional layer]
\label{def_conv}
Consider a convolutional layer with $C$ filters and $C$ input channels. In practice, each filter could have a different size, and padding is typically applied to maintain consistent output dimensions. To ease the analysis, we set all filters have the same size $(2k+1)\times(2k+1)$. Let $y_i(\alpha)\in\mathbb{R}$ represents the input to the convolutional layer with filter $i$ at position $\alpha$, then the output at position $\alpha$ can be writen as
\[
    \overline{y}_i(\alpha) = \sum\limits_{c=1}^C\sum\limits_{\beta\in ker} W_{ci,\beta}\phi(y_c(\alpha+\beta)) + b
\]
where \(ker := \{(p_0,p_1)\in\mathbb{Z}^2;|p_0|,|p_1|\leq k\}, W_{ci,\beta}\in\mathbb{R}^{C\times C}\) denotes the weight for from channel $c$ to channel $i$ at position $(\cdot+\beta)$, \(b\in\mathbb{R}^C\) is the corresponding bias term, and $\phi$ is the ReLU function.
\end{definition}

Given a mini-batch of size $N$, and a given input sequence of vectors \(X = [x_1,...,x_N] \in \mathbb{R}^{d\times N}\), batch normalization (BN) is applied as
\(\mathrm{BN}(x_i) = x_i - \mu_B/\sigma_B \odot\gamma + \beta,\)
where \(\mu_B = 1/N \sum_{i=1}^{N} x_i,\,\, \sigma_B = \sqrt{1/N \sum_{i=1}^{N} (x_i - \mu_B)^2}\). A bottleneck block of ResNet is then expressed as
\[
    \mathrm{Res}(X) = X + \mathrm{Conv}\circ\mathrm{BN}\circ\mathrm{Conv}\circ\mathrm{BN}\circ\mathrm{Conv}\circ\mathrm{BN}(X).
\]

\subsection{Mean field frameworks}

We only define the mean-field attention layer and the mean-field convolution layer here, since our goal is to evaluate the Lipschitzness of models, and the Lipschitzness of the Pre-LN Transformer and the ResNet can be calculated by simply multiplying the Lipschitz numbers of other components. 

When the input sequence length $N$ is infinitely large, it can be convenient to model self-attention as a map between probability measures~\citep{sander2022sinkformers, geshkovski2024emergence, castin2024smooth}. Indeed, the self-attention map is permutation equivalent, which enables the map from $X=[x_1,...,x_N]$ to $m(X) = \frac{1}{N}\sum_{i=1}^N \delta_{x_i}$. 

\begin{definition}[Pushforward~\citep{santambrogio2015optimal}]
\label{def_push}
For a probability measure \(\mu\) on \(\mathbb{R}^d\) and a measurable map \(\varphi: \mathbb{R}^d \to \mathbb{R}^d\), the pushforward of \(\mu\) through \(\varphi\), denoted as \(\varphi_{\#}\mu\), is the probability measure defined by \((\varphi_{\#} \mu)(B) := \mu(\varphi^{-1}(B))\) for any Borel set \(B \subset \mathbb{R}^d\), where \(\varphi^{-1}(B) := \{x \in \mathbb{R}^d : \varphi(x) \in B\}\).
\end{definition}

\begin{definition}[Mean-field self-attention~\citep{castin2024smooth}]
\label{def_mfattn}
Let \(Q, K, V \in \mathbb{R}^{d \times d}\), and define \(A := K^\top Q / \sqrt{d/M}\). Mean-field self-attention with parameters \((A, V)\) is described as:
\[
F: \mu \in \mathcal{P}_c(\mathbb{R}^d) \mapsto (\Gamma_\mu)_{\#}\mu,\quad
\Gamma_\mu(x) = \frac{\int \exp(x^\top A^\top y) V y \, d\mu(y)}{\int \exp(x^\top A^\top y) \, d\mu(y)} \quad \text{for } x \in \mathbb{R}^d.
\]
\end{definition}

Since convolution can be permutation equivariant with respect to the channels, it can also be modeled as a map between probability measures. Specifically, the convolutional layer maps the input $Y=[y_1,..., y_C]$ to $m'(Y) = \frac{1}{C}\sum_{c=1}^C\delta_{y_c}$ where $y_i(\alpha) = \sum_{\beta}W_{c,\beta}\phi(x_c(\alpha+\beta))$ is the response from channel $c$. In previous works, the number of channels is set sufficiently large to make mean field theory applicable~\citep{xiao2018dynamical}. Therefore, we can introduce the mean-field convolution based on this limit.

\begin{definition}[Mean-field convolution]
\label{def_mfconv}
Set $W\in\mathbb{R}^{C\times C\times (2k+1)^2}$. For simplicity, we denote $W_{\beta}\in\mathbb{R}$ the weight from one channel to another at position $(\cdot+\beta\mathrm)$. A mean-field convolutional layer with parameter $W$ is described as:
\[
G: \mu^\prime \in \mathcal{P}_c(\mathbb{R}) \mapsto (\Gamma_{\mu^\prime}^\prime)_{\#}\mu^\prime,\quad
\Gamma_{\mu^\prime}^\prime(y(\alpha)) = \int\sum\limits_{\beta\in ker}W_\beta y(\alpha+\beta)d\mu^\prime(Wy)+ b
\]
where $ker=\{(\gamma,\eta);(|\gamma|,|\eta|\leq k\},\,\,b \in \mathbb{R}$. Here, we ignore the Relu function to ease the analysis.
\end{definition}

\section{Topological Lipschitzness}
\label{sec_topolip}

Before defining Topological Lipschitzness, we first explain the reason why it is needed. 

Given a classifier $f_1=c_1\circ g_1$ where $g_1$ maps input data from space $X$ to $X_1$ equipped with a distance function $d_1$, and $c_1$ maps data from $(X_1,d_1)$ to $Y$, \citet{wang2016theoretical}~defines the classifier $f:X\to Y$ to be $\{\delta,\eta_1\}$-strong-robust against adversarial attacks if for any $x,x'\in X$, $\mathbb{P}(f_1(x)=f(x^\prime)|f_0(x)=f_0(x^\prime),d_0(g_0(x),g_0(x^\prime))<\delta)>1-\eta_1$. Here, $f_0$ represents an oracle that provides ground truth labels (e.g., a human annotator). Now, consider another classifier $f_2$ and assume the following relationships among the classifiers and their mappings:

\[
\begin{tikzcd}
& (X_1,d_1) \arrow[rd, "c_1"]&\\
(X,d_X) \arrow[r, "g_0"] \arrow[ru, "g_1"] \arrow[rd, "g_2"]& (X_0,d_0) \arrow[r, "c_0"] & (Y,d_Y) \\
& (X_2,d_2) \arrow[ru, "c_2"] &
\end{tikzcd}
\]

Under this framework, we can assert that $f_2$ is more robust than $f_1$ if $f_2$ is $\{\delta,\eta_2\}$-strong-robust with the same $\delta$ but a smaller $\eta_2<\eta_1$. Intuitively, this means that when the oracle considers two distinct inputs as belonging to the same class, $f_2$ has a higher possibility than $f_1$ of recognizing them as such under perturbations within $\delta$. In other words, if there exist inputs $x,x^\prime$ such that $d(x,x^\prime)<\delta$ and $d_Y(f_i(x),f_i(x^\prime))>\epsilon$ for $i=1,2$, then a more robust classifier satisfies Lip$_2<$ Lip$_1$ where Lip$_i:=\max_{x,x^\prime}d_Y(f_i(x),f_i(x^\prime))/d(x,x^\prime)\,(x,x^\prime\in X)$. Therefore, by examining the Lipschitz constants of classifiers, we can gain insights into their robustness and effectively compare different models. We transform the input and output of models into probability distributions and consider the Wasserstein distance in between to derive topological Lipschitzness.

To further analyze robustness, we transform the input and output of models into probability distributions and utilize the Wasserstein distance between them. The Wasserstein distance is a notion widely used in optimal transport, defined as a distance function between probability distributions on a given metric space. By considering the Lipschitzness of the Wasserstein distance between the input and the output (which are probability distributions) of a function, instead of considering the Lipschitz continuity between individual data points, one can investigate the global behaviors and smoothness of the function~\citep{villani2009optimal, villani2021topics}. 

However, calculating the Wasserstein distance requires that both probability distributions reside in the same dimensional space. While dimension reduction techniques can align their dimensions, this approach becomes inefficient when computing Wasserstein distances between layers of the model such as convolutional neural networks (CNNs), which typically have layers with diverse embedding dimensions, leading information loss at different scales if using dimension reduction.

To address this challenge, we introduce \textit{Topological Lipschitzness} (\textit{TopoLip}). TopoLip builds upon Wasserstein Lipschitzness and incorporates concepts from Topological Data Analysis (TDA). When comparing model robustness theoretically, their Wasserstein Lipschitz constants can be directly compared. Experimentally, the input and output are transformed into persistence diagrams, and the Wasserstein distance between these diagrams is computed. The maximum rate of change of this Wasserstein distance defines TopoLip. We will demonstrate that TopoLip is a constant multiple of the Wasserstein Lipschitz constant. 

Persistence diagrams are essential tools in TDA that capture the multi-scale topological features of data. Each feature is represented as a point in the diagram with coordinates indicating its birth and death scales. For more details on persistence diagrams and their properties, see Appendix~\ref{appendix_ph}. 

Informally, TopoLip measures the Lipschitzness of the Wasserstein distance between the persistence diagrams of a function's input and output. The relationship can be illustrated as follows:
\[
\begin{tikzcd}
\text{Input Distribution} \arrow[d ] \arrow[r, "F"] & \text{Feature Embeddings} \arrow[r, "g"] & \text{Persistence Diagrams} \\
\text{(Probability Distribution)} \arrow[ru ] &&
\end{tikzcd}
\]
Here, TopoLip combines the Lipschitzness of the function $F$ with the Lipschitz map $g$ that generates persistence diagrams. Formally, by Lemma~\ref{lem_lip_comp}, TopoLip is defined as follows:
\begin{definition}
\label{def_topolip}
    Let $g$ be a Lipschitz map defined by:
    \[
        g:\,\,\mathcal{D}\,\longrightarrow\,\mathcal{PD}_k
    \]
    \[
        g(X) = \{(b_i,d_i)|\,\,\text{feature $i$ in $H_k$ births at $b_i$ and dies at $d_i$}\,\}
    \]
    where $\mathcal{D}$ is the space of finite metric spaces (datasets), and $\mathcal{PD}_k$ is the space of persistence diagrams for dimension $k$ with the Wasserstein distance $W_p\,\,(p\geq1)$. For a Lipschitz function $F$, its Topological Lipschitzness is defined as:
    \[
        \mathrm{Lip}_{\mathrm{TopoLip}}^{W_p} (F) := \mathrm{Lip}^{W_p}(g) \cdot \mathrm{Lip}^{W_p}(F).
    \]
\end{definition}
The map $g$ is Lipschitz due to the stability theorem presented in~\citet{cohen2005stability}. When $g$ is fixed (in this work, persistent homology) to generate persistence diagrams, $\mathrm{Lip}(g)$ remains constant. Therefore, the TopoLip of a function is directly proportional to its Wasserstein Lipschitzness. By examining the Wasserstein Lipschitzness of a model, we can compare the TopoLip of different models and thus assess their robustness.

\section{Wasserstein Lipschitzness comparison}
\label{sec_wass_lip}

We begin by defining the Wasserstein Lipschitness:

\begin{definition}[Lipschitz constant with respect to the 1-Wasserstein distance \citep{castin2024smooth}]
Denote $\mathcal{P}_c(\mathbb{R}^d)$ the set of compactly supported probability measures on $\mathbb{R}^d$. P-Wasserstein distance is defined as:
\[
    W_p := \left(\inf_{\pi\in\Pi(\mu,\nu)}\int \|x - y\|^p \, d\pi(x,y)\right)^{1/p}
\]
for $\mu,\nu\in\mathcal{P}_c(\mathbb{R}^d)$, where $\Pi(\mu,\nu)$ is the set of couplings between $\mu$ and $\nu$. For a map $F:\mathcal{P}_c(\mathbb{R}^d)\to\mathcal{P}_c(\mathbb{R}^d)$ and any subset $\mathcal{X}\subset \mathcal{P}_c(\mathbb{R}^d)$, the Lipschitz constant of $F$ on $\mathcal{X}$ is defined as:
\[
    \mathrm{Lip}^{W_1}(F_{\mathcal{X}}) := \sup_{\mu, \nu \in \mathcal{X}, \mu \neq \nu} \frac{W_1(F(\mu), F(\nu))}{W_1(\mu, \nu)}.
\]
If $\mathrm{Lip}(F_{\mathcal{X}})$ is finite, then $F$ is said to be $W_2$-Lipschitz continuous on $\mathcal{P}_c(\mathbb{R}^d)$.
\end{definition}

The reason for using Lip$^{W_1}$ instead of Lip$^{W_2}$ here is because for probability measures $\mu$ and $\nu$, $W_1(\mu,\nu)\leq W_2(\mu,\nu)$ holds, meaning that the 1-Wasserstein Lipschitzness can be extended to the 2-Wasserstein Lipschitzness.

To ensure a fair comparison of variances between the self-attention and convolutional layers, we take each element of $Q, K, V, W^O$ in the self-attention layer to be drawn i.i.d. from $\mathcal{N}(0,\sigma^2)$. For the convolution layer, to follow common initialization schemes such as He initialization~\citep{he2015delving}, each element of $W$ is drawn from i.i.d. $\mathcal{N}(0,\sigma^2/(C(2k+1)^2))$. We assume $H, W, C$ in the input image size $H\times W\times C$ are very large. For the self-attention layer, the input is a sequence with size $d\times N$, where $d$ is the embedding dimension and $N=\frac{HW}{P^2}$. For the convolution layer, the input is a sequence with size $C\times N'$ where $N'=HW$.

\subsection{Attention and convolution}
\label{subsec_attn_conv}

\begin{theorem}
\label{thm_lip_attn}
Let $Q,K,V\in\mathbb{R}^{d\times d}$. For any $t>\sqrt{d}$ and $s\geq\sigma\sqrt{2\log2}$, with probability at least $\min\{1-d/t^2, 1-2e^{-s^2/(2\sigma^2)}\}$, and assuming $\|A\|_{op}\geq 2/\sigma^2$, the mean-field single-head attention map $\mathrm{Attn}_{|\mathcal{P}(B_{t\sigma})}$ with parameter $(Q,K,V)$ is $W_1$-Lipschitz continuous on the set $\mathcal{P}(B_{t\sigma})$, and its Lipschitz constant is bounded by
\[
    \mathrm{Lip}^{W_1}(\mathrm{Attn}_{|\mathcal{P}(B_{t\sigma})}) = 2t\sigma (2\sigma\sqrt{d}+s)(1+t\sigma d^{-1/2}(2\sigma\sqrt{d}+s)^2)
\]
Similarly, the Lipschitz constant of mean-field $M$-head attention map $\mathrm{MHAttn}_{|\mathcal{P}(B_{t\sigma})}$ is bounded by
\[
    \mathrm{Lip}^{W_1}(\mathrm{MHAttn}_{|\mathcal{P}(B_{t\sigma})}) = 2t\sigma\sqrt{M} (2\sigma\sqrt{d}+s)^2(1+t\sigma\sqrt{\frac{M}{d}}(2\sigma\sqrt{d}+s)^2).
\]
\end{theorem}

To simplify the upper bounds, assume $t=p\sqrt{d},\,\,s=q\sigma$ for constants $p,q>0$. Under this assumption, the Lipschitz constants of a single-head and multi-head attention layer can be approximated as follows:
\[
    \mathrm{Lip}^{W_1}(\mathrm{Attn}_{|\mathcal{P}(B_{t\sigma})}) = \mathcal{O}(\sigma^5 d^2),\,\,\mathrm{Lip}^{W_1}(\mathrm{MHAttn}_{|\mathcal{P}(B_{t\sigma})}) = \mathcal{O}(\sigma^6 d^{5/2}M).
\]

\begin{theorem}
\label{thm_lip_conv}
Let $W\in\mathbb{R}^{C\times C\times(2k+1)^2}$ where $W_{ci,\beta}\sim N(0,\frac{\sigma^2}{C(2k+1)^2})$ represents the weight from channel $c$ to channel $i$ at position $(\cdot+\beta)$. Denote the output vector of the mean-field convolutional layer as $\overline{\mathbf{y}}(\alpha) = \begin{bmatrix}\overline{y}_1(\alpha),\cdots,\overline{y}_C(\alpha)\end{bmatrix}$ where $\overline{y}_i(\alpha) = \int_{\mathbb{R}}\left(\sum_{\beta}W_{ci,\beta}y_i(\alpha+\beta)+b_i\right) d\mu(Wy)$. For any $t>0$, with probability at least $1-1/t^2$, the Lipchitz constant of the mean-field convolution map $\mathrm{Conv}_{|\mathcal{P}(B_{t\sigma})}$ with parameter $W$ is bounded by
\[
    \mathrm{Lip}^{W_1}(\mathrm{Conv}_{|\mathcal{P}(B_{t\sigma})}) = (2k+1)\sqrt{t\sigma C\left(1+\frac{1}{(2k+1)\sqrt{C}}\right)} = \mathcal{O}(k\sqrt{\sigma C}).
\]
where we assume $t$ to be some moderate positive number to simplify the upper bound.
\end{theorem}

From the above bounds, we know that the Wasserstein Lipschitzness of attention layers, as well as their TopoLip and robustness, are highly related to the embedding dimension $d$ and the head number $M$. Since $d$ and $M$ are fixed, we can indicate that Lip$^{W_1}$ of attention layers remains in a certain range. For convolution layers, since their Wasserstein Lipschitzness is related to the channel number $C$ which usually is not fixed in a model, its robustness tends to be lower than attention layers. 

Furthermore, if the bound of Lip$^{W_1}$ is tight enough, it can represent the scale or dynamics Lip$^{W_1}$. Suppose the bounds in Theorem~\ref{thm_lip_attn} and \ref{thm_lip_conv} are tight, then we can assess the Lipschitz bounds of both models from a practical perspective. In practice, typical parameter values are often set as follows: $\sigma\sim10^{-2},$ $d\sim10^2,$ $M\sim10^1,$ $k\sim10^1,$ and $C\sim10^2$. Under this setting, the Lipschitz bound for multi-head attention is on the order of $\mathcal{O}(10^{-6})$, whereas that for convolutional layers is significantly larger, around $\mathcal{O}(10^1)$. To provide a more concrete comparison, consider the following specific parameter settings: $d=512$, $M=8$, $\sigma=0.05$, $k=3$, and $C=512$. Under this setting, $\sigma^5d^2 \approx0.08,\,\,\sigma^6d^{\frac{5}{2}}M \approx0.74$, while $k\sqrt{\sigma C}\approx15$. Furthermore, it is important to note that $C$ is not fixed in practice. For instance, the number of channels in ResNet50 are 64$\to$256$\to$512$\to$1024$\to$2048, which leads to a larger Lipschitz bound for convolutional layers. Therefore, convolution is more unstable under this setting, leading to greater TopoLip and lower robustness.

Theorem~\ref{thm_lip_attn} and \ref{thm_lip_conv} indicates that while Lip$^{W_2}$ of convolution has a bound that is highly unpredictable, Lip$^{W_2}$ of attention has a fixed bound, and the bound is relatively tight under practical settings. In a real-life scenario, attention and convolution layers are rarely used solely. Instead, they are one part of the models. To conduct a thorough comparison, we extend our investigation to two widely used models: Vision Transformer (ViT) and residual neural network (ResNet).

\subsection{ViT and ResNet}
\label{subsec_vit_res}
We consider the Pre-Layer Normalized Vision Transformers (Pre-LN ViT) and ResNet. Building upon the calculations presented in Theorems~\ref{thm_lip_attn} and \ref{thm_lip_conv}, and utilizing Lemma~\ref{lem_lip_comp}, we derive the following Lipschitz constants:
\begin{align*}
    \mathrm{Lip}^{W_1}(\mathrm{TF}) &= (\mathrm{Lip}^{W_1}(\mathrm{MLP})\cdot \mathrm{Lip}^{W_1}(\mathrm{LN})+1)\cdot(1 + \mathrm{Lip}^{W_1}(\mathrm{MHAttn})\cdot \mathrm{Lip}^{W_1}(\mathrm{LN}))\\
    &= (\|W_1\|_{op}\|W_2\|_{op}\|\gamma\|_{\infty}+1)(1+\|\gamma\|_{\infty}\mathrm{Lip}^{W_1}(\mathrm{MHAttn}))\\
    &= \mathcal{O}\left( \max\left\{ 1,\sigma^3d^{3/2}M,\sigma^7d^3M,\sigma^{10}d^{9/2}M \right\} \right),\\
    \mathrm{Lip}^{W_1}(\mathrm{Res}) &= 1 + \mathrm{Lip}^{W_1}(\mathrm{Conv})^3\cdot\mathrm{Lip}^{W_1}(\mathrm{BN})^3 = \mathcal{O}\left( \max\left\{ 1,k^3\sigma^{5/2}C^3 \right\} \right).
\end{align*}
From these results, we observe that the Lipschitz constants Lip$^{W_1}$ for both ViTs and ResNets retain and further magnify the parameter dependencies inherent in their respective attention and convolutional layers. Notably, when considering the same settings as discussed in Section~\ref{subsec_attn_conv}, we find that Lip$^{W_1}(\mathrm{TF})=\mathcal{O}(1)$ for ViTs, whereas Lip$^{W_1}(\mathrm{Res})=\mathcal{O}(10^4)$ for ResNets. Additionally, since the number of channels $C$ in ResNet can be very large, the Lipschitz constant for ResNet can become significantly higher than that of ViT. As a result, ViTs tend to have a lower TopoLip value, which means they are smoother in terms of their topological properties compared to ResNets. This smoothness suggests that ViTs are less affected by changes or noise in the input, which could make them more stable and robust in their performance.

\section{Experimental results}
\label{sec_exp}

We conduct experiments using the CIFAR-10 and CIFAR-10C dataset to evaluate the relationship between TopoLip and robustness. Specifically, we train ResNet18/50/101 and three ViTs (small, base, and large) for practical settings. We also train two convolution-only models (Conv) and two attention-only models (Attn), each with small and large configurations, to verify the theoretical results for attention and convolution layers.

For Convs, the small configuration uses up to 64 channels across all layers, while the large configuration scales up to 2048 channels in the final layers. For Attn models, the small version features 4 attention heads with an embedding dimension of 128, whereas the large version uses 12 heads with an embedding dimension of 512. All Convs and Attns have 10 layers. Detailed configurations for Attn, Conv, ResNet, and ViT architectures are provided in Table~\ref{tbl_config}.

We train Attns for 100 epochs, Convs for 200 epochs, ResNets for 100 epochs, and ViTs for 200 epochs on CIFAR10 for each model to reach optimal or near-optimal performance levels under the given configurations. ResNet models achieved validation accuracies exceeding 90\%, while ViTs range from 77.8\% to 87.0\% (Figure~\ref{fig_vit_res_acc}). Attn models, however, showed much lower validation accuracies, with both configurations remaining below 35\%, reflecting the limitations of their simplified architectures. In contrast, Conv models performed significantly better, with the small configuration achieving 57.1\% and the large configuration reaching 85.7\%, despite their simple designs (Figure~\ref{fig_attn_conv_acc}). From the loss curves, we observed that only the training of the large Attn model failed under its simple configuration. This could be attributed to the behavior of attention layers in the early training stages, where they amplify the importance of certain positions or data points. Once a position is deemed important, its attention score increases, reinforcing its significance as training progresses. Without mechanisms like layer normalization to mitigate this effect, the training process can converge prematurely, hindering further weight updates. In fact, the large Attn model's loss failed to record after the first epoch, causing its loss curve in Figure~\ref{fig_attn_conv_acc} to appear ``lost''.

\begin{table}[t]
\caption{Model configurations.}
\label{tbl_config}
\begin{center}
\begin{tabular}{ll}
\multicolumn{1}{c}{\bf Model} & \multicolumn{1}{c}{\bf Configuration} \\ \hline \\
Attn (small) & 4 heads; embedding dimension: 128\quad  (h4 d128)\\
Attn (large) & 12 heads; embedding dimension: 512\quad  (h12 d512)\\
Conv (small) & $\#$(channel): 3$\to$64$\to$64$\to$64$\to$64$\to$64$\to$64$\to$64$\to$64$\to$64 \\
Conv (large) & $\#$(channel): 3$\to$64$\to$64$\to$128$\to$128$\to$256$\to$512$\to$1024$\to$2048$\to$2048\\
ViT (small) & 6 heads; embedding dimension: 384\quad  (h6 d384)\\
ViT (base) & 12 heads; embedding dimension: 768\quad  (h12 d768)\\
ViT (large) & 16 heads; embedding dimension: 1024\quad  (h16 d1024)\\
ResNet18 & $\#$(channel): 3$\to$64$\to$64($\times$2)$\to$128($\times$2)$\to$256($\times$2)$\to$512($\times$2)\\
ResNet50 & $\#$(channel): 3$\to$64$\to$64($\times$3)$\to$128($\times$4)$\to$256($\times$6)$\to$512($\times$3)\\
ResNet101 & $\#$(channel): 3$\to$64$\to$64($\times$3)$\to$128($\times$4)$\to$256($\times$23)$\to$512($\times$3)
\end{tabular}
\end{center}
\end{table}

Next, we evaluate the TopoLip of the models to understand their robustness. To measure the Wasserstein distance between the persistence diagrams of the input and output at each layer (or each block for ResNets), we first switch the models to evaluation mode to freeze their parameters. Then, we input the test dataset and collect the outputs from all layers. Using these outputs, we compute their persistence diagrams and calculate the Wasserstein distances between adjacent layers. Next, we compute the absolute change rates. If the Wasserstein distances of two adjacent layers are $WD_1$ and $WD_2$, the absolute change rate between them is defined as $|(WD_2-WD_1)/WD_1|$. Finally, the TopoLip is computed as the \textbf{maximum absolute change rate} observed across all layers (Table~\ref{tbl_topolip}). 

\begin{table}[ht!]
\centering
\caption{Topolip on CIFAR-10 and CIFAR-10C Corruptions (\%). Abbreviations: "s, b, l" stand for small, base, and large, respectively; "Res" stands for ResNet; $\mathbf{W_i}$ stands for the $i$-Wasserstein distance. \textbf{Highlighted} values indicate relatively lower results in comparisons of Attn vs. Conv and ViT vs. ResNet for $\mathbf{W_1}$ and $\mathbf{W_2}$, respectively.}
\label{tbl_topolip}
\begin{tabular}{@{}l@{\hskip 3pt}c@{\hskip 3pt}c@{\hskip 3pt}c@{\hskip 3pt}c@{\hskip 3pt}c@{\hskip 3pt}c@{\hskip 3pt}c@{\hskip 3pt}c@{\hskip 3pt}c@{\hskip 3pt}c@{}}
\toprule
& \textbf{Attn(s)} & \textbf{Attn(l)} & \textbf{Conv(s)} & \textbf{Conv(l)} & \textbf{ViT(s)} & \textbf{ViT(b)} & \textbf{ViT(l)} & \textbf{Res18} & \textbf{Res50} & \textbf{Res101} \\ \midrule
$\mathbf{W_1}(\downarrow)$ &\textbf{0.57} & \textbf{1.00} & 6.83 & 16.20 & \textbf{0.97} & \textbf{1.15} & \textbf{0.82} & 2.29 & 1.45 & 1.50\\
$\mathbf{W_2}(\downarrow)$ &\textbf{0.52} & \textbf{1.00} & 4.13 & 7.27 & \textbf{0.56} & \textbf{0.64} & \textbf{0.46} & 1.10 & \textbf{0.60} & 1.02\\
 \bottomrule
\end{tabular}
\end{table}

From Table~\ref{tbl_topolip}, we know that Attns have relatively lower Wasserstein Lipschitz constants than Convs, and ViTs have lower Wasserstein Lipschitz constants than ResNets, which align with the theoretical results. While TopoLip provides certain insights into the robustness of models, we propose that rather than focusing solely on TopoLip, analyzing the entire change rate landscape offers a deeper understanding of the model's robustness.

\begin{figure}[ht]
    \centering
    \includegraphics[width=0.8\linewidth]{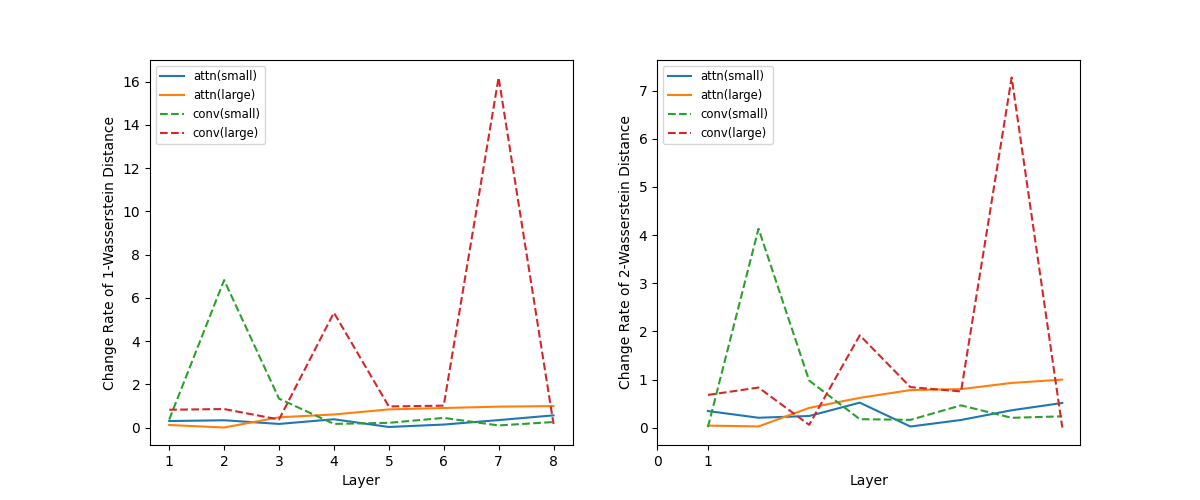}
    \caption{Absolute change rate of the Wasserstein distance of persistence diagrams of Attns and Convs.}
    \label{fig_attn_conv_ddist}
\end{figure}

\begin{figure}[ht]
    \centering
    \includegraphics[width=0.8\linewidth]{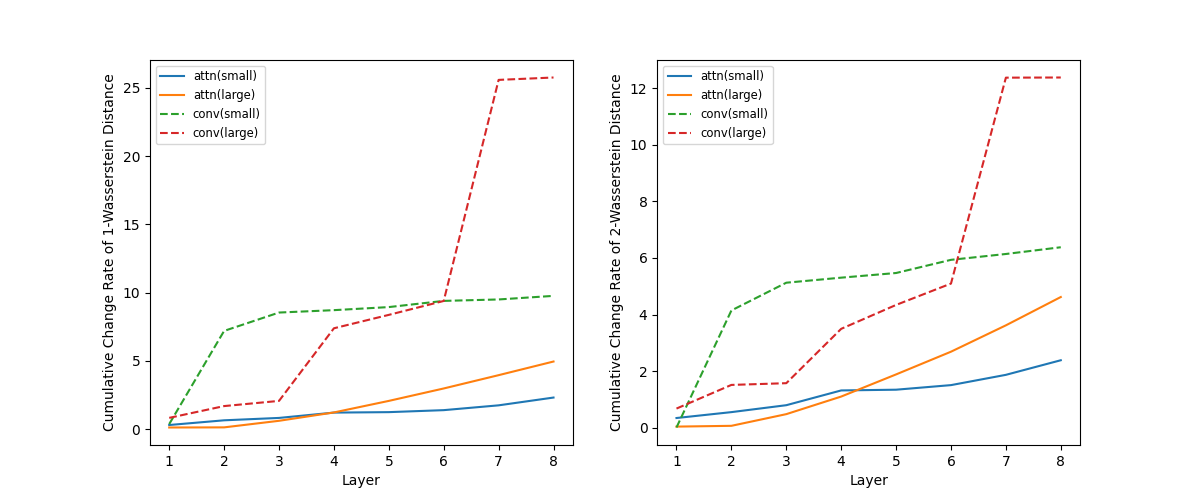}
    \caption{Cumulative absolute change rate of the Wasserstein distance of persistence diagrams of Attns and Convs.}
    \label{fig_attn_conv_cuddist}
\end{figure}

\begin{figure}[ht]
    \centering
    \includegraphics[width=0.8\linewidth]{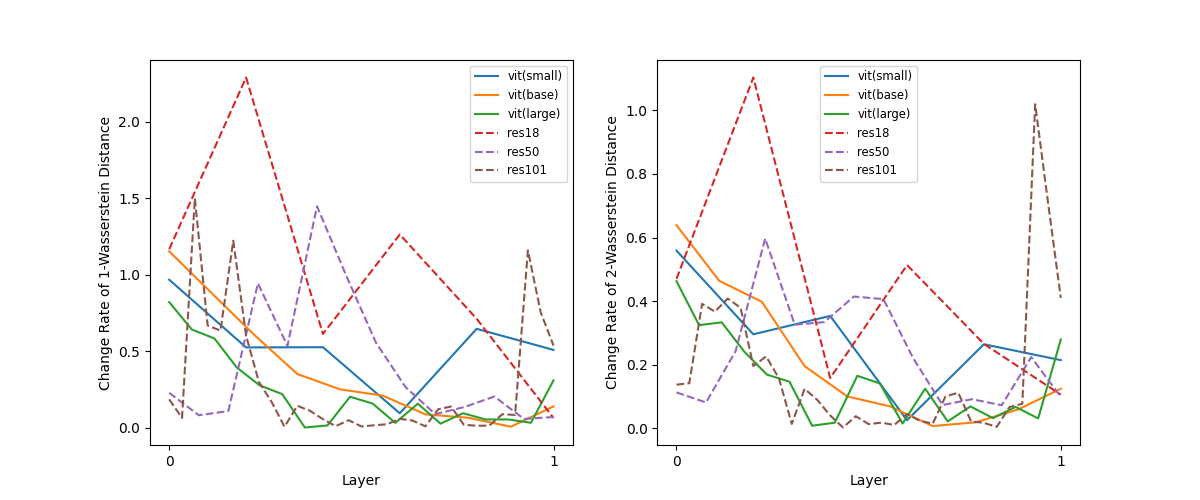}
    \caption{Absolute change rate of the Wasserstein distance of persistence diagrams of ViTs and ResNets.}
    \label{fig_vit_res_ddist}
\end{figure}

\begin{figure}[ht]
    \centering
    \includegraphics[width=0.8\linewidth]{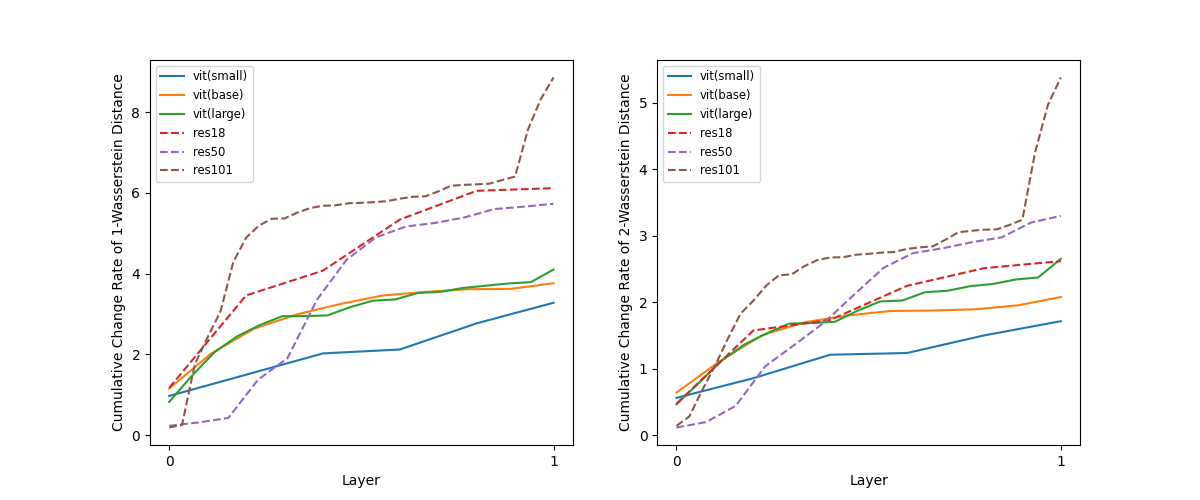}
    \caption{Cumulative absolute change rate of the Wasserstein distance of persistence diagrams of ViTs and ResNets.}
    \label{fig_vit_res_cuddist}
\end{figure}

The results of the absolute change rate and cumulative absolute change rate are shown in Figure~\ref{fig_attn_conv_ddist} to \ref{fig_vit_res_cuddist}. Since ResNets and ViTs have different numbers of layers, we interpolate their results to align them on a normalized scale from layer 0 to 1 for consistent comparison. From Figures~\ref{fig_attn_conv_ddist} and \ref{fig_attn_conv_cuddist}, we observe that Convs exhibit higher maximum change rates (TopoLip) compared to Attns, indicating that Attns are more robust and topologically smooth than Convs, which aligns with the theoretical results in Section~\ref{subsec_attn_conv}. From Figure~\ref{fig_vit_res_ddist}, we see that ResNet models have a higher TopoLip than ViTs, with their change rates displaying more turbulent behavior. This is visualized more clearly in Figure~\ref{fig_vit_res_cuddist}. Interestingly, the cumulative 2-Wasserstein change rate of the ViT (large) model is comparable to that of ResNet18, suggesting that they share a similar level of robustness. Moreover, the cumulative 2-Wasserstein curve for ViT (base) progresses slightly slower but remains close to those of ViT (large) and ResNet18. This indicates comparable robustness levels, with ViT (base) showing a marginally higher robustness. The corresponding Wasserstein distances are shown in Figure~\ref{fig_vit_res_dist} and \ref{fig_attn_conv_dist}.

Finally, we evaluate the robustness of the models using the CIFAR-10C dataset. CIFAR-10C is an extended version of CIFAR-10, designed to assess model robustness by introducing 15 types of common corruptions, each applied at five severity levels. For our evaluation, we focus on five corruption types: Gaussian Noise, Motion Blur, Snow, Impulse Noise, and Pixelate. The results are summarized in Table~\ref{tbl_performance}, where models demonstrating superior robustness (compared to their baselines) are \textbf{highlighted}.

From Table~\ref{tbl_performance}, we observe that ViTs (small and base) are generally more robust than ResNets across the selected corruption tasks. However, ViT (large) does not exhibit the same level of robustness as its smaller counterparts. Instead, its robustness appears closer to that of ResNet18. Moreover, the robustness of ViT (base) is relatively close to ResNet18 compared to other ResNets. These results align with the earlier analysis.

For Attn and Conv models, while Attn models demonstrate significant robustness, we hypothesize that this is not solely due to the architecture itself. Instead, the limited training capacity of Attn models in such simple configurations likely results in low baseline performance, which can make even slight performance improvements appear substantial in comparison. Overall, the robustness results align with the Wasserstein change rate findings, underscoring the close relationship between TopoLip and model robustness. Furthermore, analyzing the whole Wasserstein distance change rate landscape along with TopoLip provides a more comprehensive understanding of a model's robustness characteristics.

\begin{table}[h!]
\centering
\caption{Model Performance on CIFAR-10 and CIFAR-10C Corruptions (\%).}
\label{tbl_performance}
\begin{tabular}{@{}l@{\hskip 5pt}c@{\hskip 5pt}c@{\hskip 5pt}c@{\hskip 5pt}c@{\hskip 5pt}c@{\hskip 5pt}c@{}}
\toprule
\textbf{Model}   & \textbf{CIFAR-10} & \textbf{Gauss Noise} & \textbf{Motion Blur} & \textbf{Snow} & \textbf{Impulse Noise} & \textbf{Pixelate} \\ \midrule
Attn (small) & 34.2 & \textbf{36.5 (+2.3)} & \textbf{32.5 (-1.7)} & \textbf{30.0 (-4.2)} & \textbf{35.2 (+1.0)} & \textbf{35.2 (+1.0)} \\

Attn (large) & 12.0 & \textbf{16.9 (+4.9)} & \textbf{17.2 (+5.2)} & \textbf{14.8 (+2.8)} & \textbf{16.9 (+4.9)} & \textbf{17.0 (+5.0)} \\

Conv (small) & 57.1 & 31.3 (-25.8) & 36.5 (-20.6) & 38.0 (-19.1) & 22.6 (-34.5) & \textbf{42.7 (-14.4)} \\

Conv (large) & 85.7 & 44.4 (-41.3) & 51.2 (-34.5) & 61.6 (-24.1) & 23.8 (-61.9) & 60.8 (-24.9) \\

ViT (small) & 77.8 & \textbf{53.4 (-24.4)} & \textbf{54.0 (-23.8)} & \textbf{55.3 (-22.5)} & 39.1 (-38.7) & \textbf{62.7 (-15.1)} \\

ViT (base) & 85.2 & \textbf{60.8 (-24.4)} & \textbf{69.9 (-15.3)} & \textbf{75.8 (-9.4)} & 42.2 (-43.0) & \textbf{78.3 (-6.9)} \\

ViT (large)  & 87.0  & 41.4 (-45.6) & 39.8 (-47.2) & 36.1 (-50.9) & 40.3 (-46.7) & 41.7 (-45.3) \\

ResNet18  & 90.9 & 50.0 (-40.9) & \textbf{63.0 (-27.9)} & \textbf{73.3 (-17.6)} & 32.8 (-58.1) & 56.1 (-34.8) \\

ResNet50  & 91.4 & 50.0 (-41.4) & 60.3 (-31.1) & \textbf{73.1 (-18.3)} & 33.9 (-57.5) & 54.8 (-36.6) \\

ResNet101 & 91.8 & 51.9 (-39.9) & 61.2 (-30.6) & \textbf{75.0 (-16.8)} & 35.1 (-56.7) & 66.6 (-25.2) \\ \bottomrule
\end{tabular}
\end{table}

\section{Conclusion}

\paragraph{Summary.} In this paper, we introduced \textit{TopoLip}, a metric designed to assess the robustness of machine learning models at a layer-wise level. TopoLip is effective for both theoretical and experimental comparisons of different architectures or configurations, and can provide insights into how model robustness depends on parameters. Through theoretical analysis of the Wasserstein-Lipschitz conditions in mean-field attention and convolution, we revealed that attention-based models are inherently smoother than convolutional models, making them more robust as defined by TopoLip. Experimental results further validated these findings, showing that attention-based models exhibit greater robustness than convolution-based models when handling corrupted data.

\paragraph{Limitations.} Our theoretical analysis is limited to providing upper bounds for the Lipschitz constants of relatively simple models. While these bounds align with empirical results, extending the analysis to more complex architectures and incorporating lower bounds would enhance its rigor and applicability.

\textit{Despite these limitations, our work establishes a foundation for bridging theoretical and empirical analyses of robustness. To our knowledge, TopoLip is the first metric to enable direct robustness comparisons across different architectures or configurations through both theory and experiments.}

\bibliographystyle{plainnat}

\clearpage
\appendix
\section{Persistence homology}
\label{appendix_ph}
\begin{figure}[H]
\begin{center}
    \includegraphics[width = 0.5\linewidth]{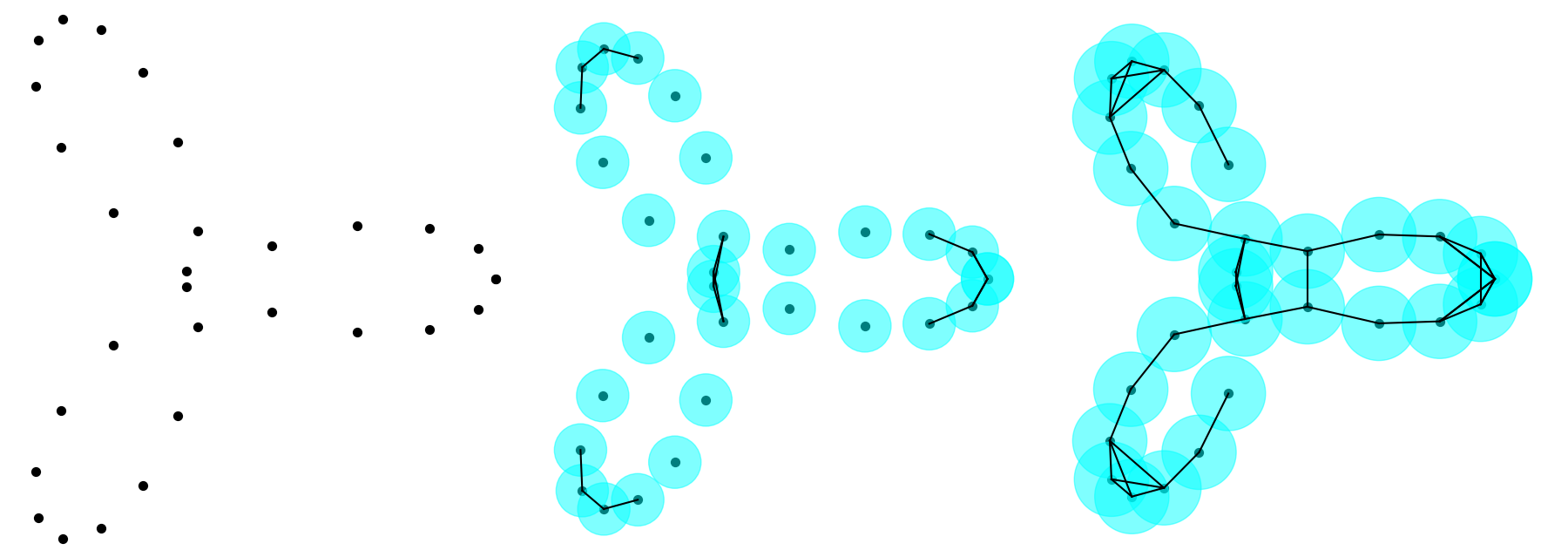}
\end{center}
\caption{Contruction of the \v{C}ech complex of the dataset.}
\label{fig_filtration}
\end{figure}

We provide an intuitive overview of persistence homology, omitting a formal introduction that can be found in~\citep{le2018persistence, bubenik2015statistical, naitzat2020topology}. Filtration is a key technique in capturing the topological features of data. Among various types of filtrations, the \v{C}ech complex is widely used. The \v{C}ech complex constructs a topological structure by forming simplices based on the intersections of balls with a specific radius centered at each data point (Figure~\ref{fig_filtration}). As the radius increases, more simplices are added, allowing the complex to capture topological features at different scales.

\begin{figure}[H]
\begin{center}
    \includegraphics[width = 0.38\linewidth]{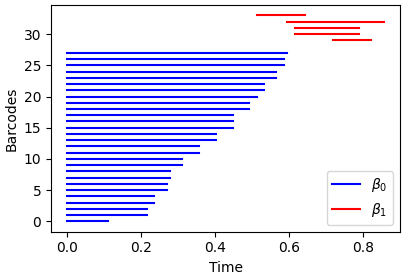}
    \includegraphics[width = 0.28\linewidth]{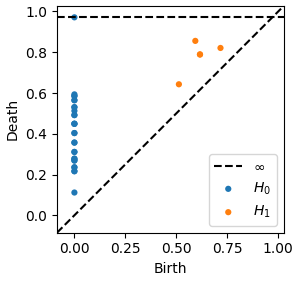}
\end{center}
\caption{Persistence barcode and persistence diagram.}
\label{fig_pers}
\end{figure}

During the filtration process of the \v{C}ech complex, topological features such as connected components and holes emerge and disappear. These events are recorded using persistence barcodes, which track the birth and death of each feature (Figure~\ref{fig_pers}). Here, $\beta_0$ and $\beta_1$ represent the lifespan of connected components and 2D holes, respectively. The barcodes are then represented as points in a persistence diagram, which is a multiset of points in the Cartesian plane $\mathbb{R}^2$. In the persistence diagram, $H_0$ corresponds to connected components and $H_1$ to 2D holes. Since the persistence diagram Dg can be considered as a discrete measure $\mu_{Dg} = \sum_{u\in Dg}\delta_u$ where $\delta_u$ is the Dirac unit mass on $u$, the bottleneck distance is usually used to measure the difference between persistence diagrams \citep{le2018persistence, adams2017persistence}. Additionally, 1- and 2-Wasserstein distances are also frequently used \citep{berwald2018computing}.

\begin{lemma}[(Lipschitz Constant of Composed Functions~\citep{gouk2021regularisation})]
\label{lem_lip_comp}
    Let $(X, d_X)$, $(Y, d_Y)$, and $(Z, d_Z)$ be metric spaces. Suppose that $f: X \to Y$ is Lipschitz continuous with Lipschitz constant $L_f$, and $g: Y \to Z$ is Lipschitz continuous with Lipschitz constant $L_g$. Then the composition $g \circ f: X \to Z$ is Lipschitz continuous with Lipschitz constant at most $L_f \cdot L_g$. In other words, for all $x_1, x_2 \in X$,
\[
d_Z(g(f(x_1)), g(f(x_2))) \leq L_f \cdot L_g \cdot d_X(x_1, x_2).
\]
\end{lemma}

\section{Proof of section 3}
\begin{lemma}[\citep{vershynin2010introduction}]
\label{lem_singular}
\textit{Given a matrix $A\in\mathbb{R}^{d\times d}$ with entries $A_{ij}\sim_{i.i.d.} N(0,\sigma^2)$, denote the singular values as $s_1(A)\geq s_2(A)\geq \cdots\geq s_d(A)\geq0$. Then:}
\[
    P[s_1\leq 2\sigma\sqrt{d}+t]\geq 1-2e^{-\frac{t^2}{2\sigma^2}}.
\]
\end{lemma}

\paragraph{Proof of Theorem \ref{thm_lip_attn}.}
We begin by bounding the Lipschitz constant for single-head attention. While \citet{castin2024smooth} provides an upper bound for $\mathrm{Lip}(\mathrm{Attn}_{|\mathcal{P}(B_{t\sigma})})$, their proof is abbreviated. Here, we present the comprehensive proof and offer a potentially tighter lower bound. We also extend the analysis to multi-head attention by providing an upper bound for $\mathrm{Lip}(\mathrm{MHAttn}_{|\mathcal{P}(B_{t\sigma})})$. 

Define the kernel function $K(x,y) := e^{x^\top A^\top y}$. The mean-field attention map is then expressed as:
\[
    \Gamma_\mu (x) = \int_{\mathbb{R}^d}\frac{K(x,y)Vy}{\int K(x,y)d\mu(y)}d\mu(y).
\]
To bound the Lipschitz constant, we consider the difference between $\Gamma_\mu$ and $\Gamma_\nu$ for two probability measures $\mu$ and $\nu$ in $\mathcal{P}(B_{t\sigma})$:
\begin{align*}
    \|\Gamma_\mu (x)& - \Gamma_\nu (x)\|_{L^\infty (B_{t\sigma},\mathbb{R}^d)} \\
    &= \biggl|\frac{\int_{\mathbb{R}^d} K(x,y)Vyd\mu(y)\int_{\mathbb{R}^d} K(x,y)d\nu(y) - \int_{\mathbb{R}^d} K(x,y)Vyd\nu(y)\int_{\mathbb{R}^d} K(x,y)d\mu(y)}{\int_{\mathbb{R}^d} K(x,y)d\mu(y)\int_{\mathbb{R}^d} K(x,y)d\nu(y)}\biggr|.
\end{align*}

Denote $y^*:=\max\limits_{y\in B_{t\sigma}}\|y\|$. We bound the numerator first:
\begin{align*}
    &\biggl|\int_{\mathbb{R}^d} K(x,y)Vyd\mu(y)\int_{\mathbb{R}^d} K(x,y)d\nu(y) - \int_{\mathbb{R}^d} K(x,y)Vyd\nu(y)\int_{\mathbb{R}^d} K(x,y)d\mu(y)\biggr|\\
    &= \biggl|\int_{\mathbb{R}^d} K(x,y)Vyd\mu(y)\int_{\mathbb{R}^d} K(x,y)(d\nu-d\mu)(y) \\
    &\qquad\qquad\qquad\qquad\qquad\qquad\qquad- \int_{\mathbb{R}^d} K(x,y)Vy(d\nu-d\mu)(y)\int_{\mathbb{R}^d} K(x,y)d\mu(y)\biggr|\\
    &\leq \biggl|\int_{\mathbb{R}^d} K(x,y)d\mu(y)\biggr|
    \biggl(\|V\|_{op}y^*\biggl|\int_{\mathbb{R}^d} K(x,y)(d\nu-d\mu)(y)\biggr| + \biggl|\int_{\mathbb{R}^d} K(x,y)Vy(d\nu-d\mu)(y)\biggr|  \biggr)\\
    &\leq 2\|V\|_{op}y^*\biggl|\int_{\mathbb{R}^d} K(x,y)d\mu(y)\biggr|
    \biggl|\int_{\mathbb{R}^d} K(x,y)(d\nu-d\mu)(y)\biggr|\\
    &\leq 2\|V\|_{op}y^* \biggl|\int_{\mathbb{R}^d} K(x,y)d\mu(y)\biggr|
    \|K(x,\cdot)\|_{C^{0,1}(B_{t\sigma})}W_1(\mu,\nu)\\
    &\leq 2y^* \|V\|_{op}\biggl|\int_{\mathbb{R}^d} K(x,y)d\mu(y)\biggr|\|K(x,\cdot)\|_{C^{0,1}(B_{t\sigma})} W_2(\mu,\nu)
\end{align*}
where we use the inequality $W_1(\mu,\nu) \leq W_2(\mu,\nu)$. By Lemma \ref{lem_singular}, with probability at least $1-2e^{-s^2/(2\sigma^2)}$, we have $\|V\|_{op}\leq 2\sigma\sqrt{d}+s$, $\|A\|_{op} \leq \sqrt{\frac{M}{d}}\|K\|_{op}\|Q\|_{op}\leq \sqrt{\frac{M}{d}}(2\sigma\sqrt{d}+s)^2$. For $\|K(x,\cdot)\|_{C^{0,1}(B_{t\sigma})}$, we can bound it as follows:
\begin{align*}
    &\|K(x,\cdot)\|_{C^{0,1}(B_{t\sigma})} \\
    &= \sup\limits_{y\in B_{t\sigma}}|K(x,y)| + \sup\limits_{y_1\neq y_2\in B(0,t\sigma)}\frac{|K(x,y_1)-K(x,y_2)|}{\|y_1-y_2\|}\\
    &\leq \sup\limits_{y\in B_{t\sigma}}|K(x,y)| + \sup\limits_{y\in B_{t\sigma}}\|\nabla_y K(x,y)\|\\
    &\leq K^*(x,y) + y^*\|A\|_{op}K^*(x,y)\\
    &= K^*(x,y)(1+y^*\|A\|_{op})
\end{align*}

where $K^*(x,y) := \sup_{y\in B_{t\sigma}}K(x,y) = \exp(y^*\|x^\top A\|)$ and the first inequality follows from the definition of the $C^{0,1}$ norm and the mean value theorem. Then $\|\Gamma_\mu (x) - \Gamma_\nu (x)\|_{L^\infty (B_{t\sigma},\mathbb{R}^d)}$ can be bounded by 
\begin{align*}
    &\|\Gamma_\mu (x) - \Gamma_\nu (x)\|_{L^\infty (B_{t\sigma},\mathbb{R}^d)} \\
    &\leq \frac{2y^* \|V\|_{op}\biggl|\int_{\mathbb{R}^d} K(x,y)d\mu(y)\biggr|K^*(x,y)(1+y^*\|A\|_{op})}{\biggl|\int_{\mathbb{R}^d} K(x,y)d\mu(y)\int_{\mathbb{R}^d} K(x,y)d\nu(y)\biggr|}W_2(\mu,\nu)\\
    &= 2y^* \|V\|_{op}(1+y^*\|A\|_{op})\frac{K^*(x,y)}{\int_{\mathbb{R}^d} K(x,y)d\nu(y)}W_2(\mu,\nu).
\end{align*}
To bound the integral part, we transform $\int d\nu(y)$ to $\int p(y)dy$ where $p(y)$ is the probability density function (pdf) of $y$. Since $y\sim N(0,\sigma^2I)$, by using the pdf of the multivariate Gaussian distribution, we have
\begin{align*}
    \int_{R^{d}} K(x,y) d\nu(y) &= \int_{R^{d}} K(x,y)p(y) dy\\
    &= \frac{1}{(2\pi\sigma^2)^{d/2}}\int_{R^{d}} e^{x^\top Ay}\cdot e^{-\|y\|^2/(2\sigma^2)} d y\\
    &= e^{\sigma^2\|x^\top A\|^2/2}\frac{1}{(2\pi\sigma^2)^{d/2}}\int_{R^{d}} e^{-\|y - \sigma^2x^\top A\|^2/(2\sigma^2)} d y\\
    &= e^{\sigma^2\|x^\top A\|^2/2}.
\end{align*}
Therefore, 
\[
    \frac{K^*(x,y)}{\int_{\mathbb{R}^d} K(x,y)d\nu(y)} = \exp(y^*\|x^\top A\| - \sigma^2\|x^\top A\|^2/2).
\]
To bound it at $1$, we need to ensure that
\[
    y^* \leq \frac{\sigma^2}{2}\|x^\top A\| \leq \frac{y^*\sigma^2}{2}\| A\|_{op} \,\,\Longrightarrow\,\, \|A\|_{op} \geq \frac{2}{\sigma^2}.
\]
holds. Under this condition, the final bound is
\[
    \|\Gamma_\mu (x) - \Gamma_\nu (x)\|_{L^\infty (B_{t\sigma},\mathbb{R}^d)}
    \leq 2y^* \|V\|_{op}(1+y^*\|A\|_{op})W_2(\mu,\nu)=: \text{Lip}(\text{Attn})W_2(\mu,\nu).
\]

Finally, since
\[
    \Gamma_\mu^{\text{MHAttn}} (x) - \Gamma_\nu^{\text{MHAttn}} (x) = W^O\begin{bmatrix}
        \Gamma_\mu^{1} (x) - \Gamma_\nu^{1} (x)\\
        \vdots\\
        \Gamma_\mu^{M} (x) - \Gamma_\nu^{M} (x)
    \end{bmatrix}
\]
where $\Gamma_\nu^{k} (x)$ denotes the mean-field self-attention of $k$-th head, we have
\begin{align*}
    &\|\Gamma_\mu^{\text{MHAttn}} (x) - \Gamma_\nu^{\text{MHAttn}} (x)\|_{L^\infty (B_{t\sigma},\mathbb{R}^d)}\\
    &\leq \|W^O\|_{op}\Biggl\|\begin{bmatrix}
        \Gamma_\mu^{1} (x) - \Gamma_\nu^{1} (x)\\
        \vdots\\
        \Gamma_\mu^{M} (x) - \Gamma_\nu^{M} (x)
    \end{bmatrix}\Biggr\| \\
    &\leq \|W^O\|_{op}\sqrt{\sum\limits_{i=1}^M\text{Lip}(\text{Attn}_{|\mathcal{P}(B_{t\sigma})})^2}\\
    &\leq 2y^* \sqrt{M}\|W^O\|_{op}\|V\|_{op}(1+y^*\|A\|_{op})W_2(\mu,\nu)=:\text{Lip}(\text{MHAttn})W_2(\mu,\nu).
\end{align*}
With probability at least $\min\{1-d/t^2, 1-2\exp(-s^2/(2\sigma^2))\}$, we can bound the terms by $y^*=t\sigma,\,\,\|W^O\|_{op},\|V\|_{op}\leq 2\sigma\sqrt{d}+s,\,\,\|A\|_{op}\leq \sqrt{M/d}\|K\|_{op}\|Q\|_{op}\leq \sqrt{M/d}(2\sigma\sqrt{d}+s)^2$. Therefore, the final bounds become
\[
    \|\Gamma_\mu (x) - \Gamma_\nu (x)\|_{L^\infty (B_{t\sigma},\mathbb{R}^d)}
    \leq 2t\sigma (2\sigma\sqrt{d}+s)(1+t\sigma d^{-1/2}(2\sigma\sqrt{d}+s)^2)W_2(\mu,\nu),
\]
\[
    \|\Gamma_\mu^{\text{MHAttn}} (x) - \Gamma_\nu^{\text{MHAttn}} (x)\|_{L^\infty (B_{t\sigma},\mathbb{R}^d)}
    \leq 2t\sigma\sqrt{M} (2\sigma\sqrt{d}+s)^2(1+t\sigma\sqrt{\frac{M}{d}}(2\sigma\sqrt{d}+s)^2)W_2(\mu,\nu)
\]
where $M=1$ for the single-head attention.
\qed

\paragraph{Proof of Theorem \ref{thm_lip_conv}.}
We begin by bounding the Lipschitz constant for a single response $\overline{y}(\alpha)$. We denote $\overline{y}^\mu(\alpha) = \int_{\mathbb{R}}\left(\sum_{\beta}W_{\beta}y_i(\alpha+\beta)+b_i\right) d\mu(Wy)$, then
\begin{align*}
    &|\overline{y}^\mu(\alpha) -  \overline{y}^\nu(\alpha)|\\
    &= \left| \int_{\mathbb{R}}\left(\sum_{\beta}W_{\beta}y(\alpha+\beta)+b_i\right) d\mu(Wy) - \int_{\mathbb{R}}\left(\sum_{\beta}W_{\beta}y(\alpha+\beta)+b_i\right) d\nu(Wy) \right| \\
    &= \left| \int_{\mathbb{R}}\left(\sum_{\beta}W_{\beta}y(\alpha+\beta)+b_i\right) (d\mu-d\nu)(Wy) \right| \\
    &\leq \Biggl\|\left(\nabla_{W}\left(\sum_{\beta}W_{\beta}y(\alpha+\beta)+b_i\right),\nabla_{y}\left(\sum_{\beta}W_{\beta}y(\alpha+\beta)+b_i\right)\right)\Biggr\|_2 W_1(\mu,\nu)\\
    &\leq\sqrt{|\sum_{\beta}y(\alpha+\beta)| + |\sum_{\beta}W_\beta|}\,\, W_1(\mu,\nu)\\
    &\leq \sqrt{\sum_{\beta}|y(\alpha+\beta)| + \sum_{\beta}|W_\beta|}\,\, W_1(\mu,\nu)\\
    &\leq (2k+1)\sqrt{t\sigma + \frac{t\sigma}{(2k+1)\sqrt{C}}}\,\,W_2(\mu,\nu) =: L\,W_2(\mu,\nu).
\end{align*}

Finally, since $\Gamma_{\mu}^\prime(\alpha) = \overline{\mathbf{y}}(\alpha)$, we can bound the difference between $\Gamma_{\mu}^\prime$ and $\Gamma_{\nu}^\prime$ as:
\begin{align*}
    \|\Gamma_\mu^\prime (\alpha) - \Gamma_\nu^\prime (x)\|_{L^\infty (B_{t\sigma},\mathbb{R}^d)}
    &= \sqrt{\sum\limits_{i=1}^C|\overline{y}^\mu(\alpha) -  \overline{y}^\nu(\alpha)|^2}\\
    &\leq \sqrt{C}L\,W_2(\mu,\nu)\\
    &= (2k+1)\sqrt{t\sigma C\left(1+\frac{1}{(2k+1)\sqrt{C}}\right)}\,\,W_2(\mu,\nu).
\end{align*}
\qed

\section{Further experimental results}
\label{appendix_exp}

\begin{figure}
    \centering
    \includegraphics[width=0.7\linewidth]{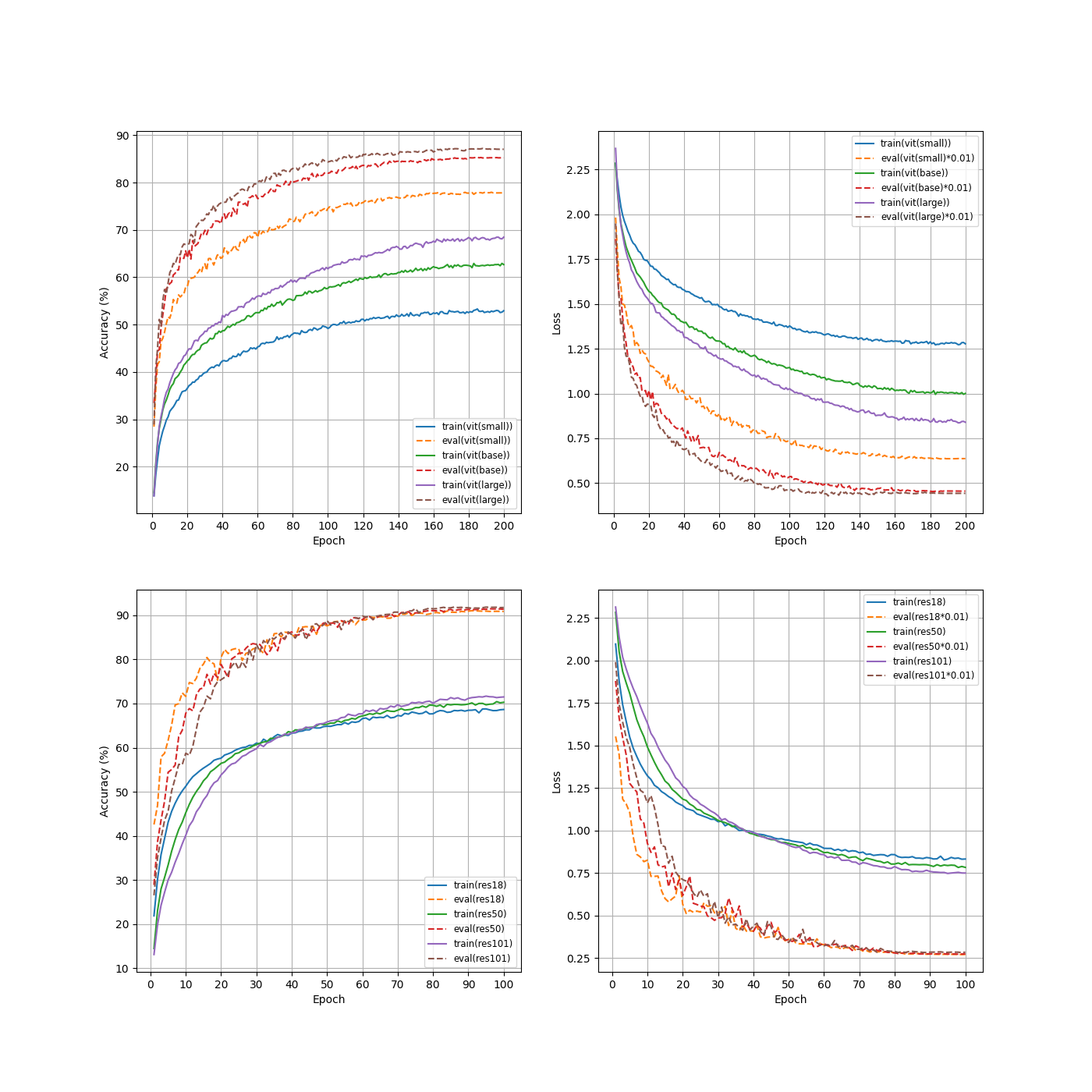}
    \caption{Accuracy and loss of ViTs and ResNets.}
    \label{fig_vit_res_acc}
\end{figure}

\begin{figure}
    \centering
    \includegraphics[width=0.7\linewidth]{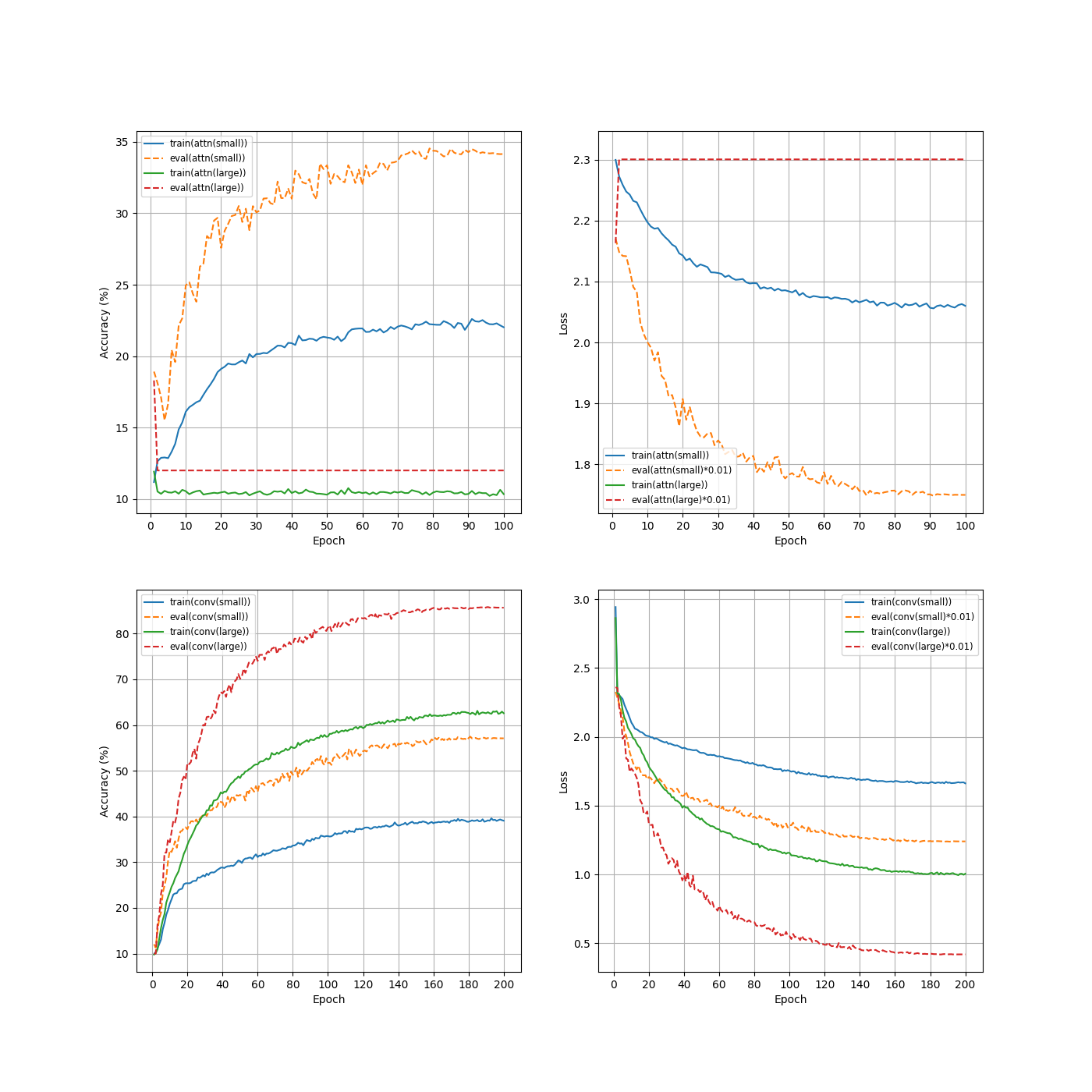}
    \caption{Accuracy and loss of Attns and Convs.}
    \label{fig_attn_conv_acc}
\end{figure}

\begin{figure}
    \centering
    \includegraphics[width=0.8\linewidth]{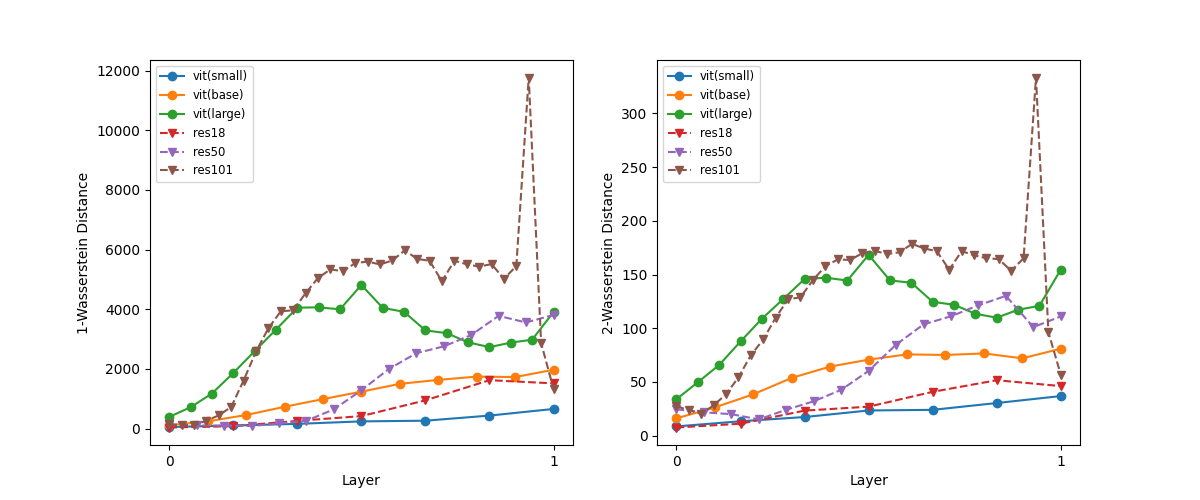}
    \caption{Wasserstein distance of the persistence diagrams of ViTs and ResNets.}
    \label{fig_vit_res_dist}
\end{figure}

\begin{figure}
    \centering
    \includegraphics[width=0.8\linewidth]{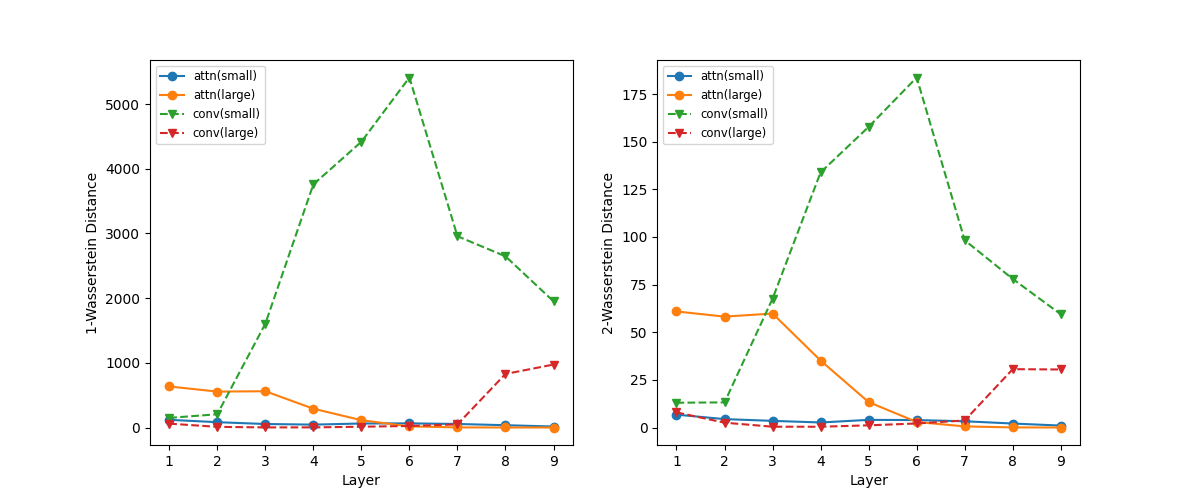}
    \caption{Wasserstein distance of the persistence diagrams of Attns and Convs.}
    \label{fig_attn_conv_dist}
\end{figure}

Figure~\ref{fig_vit_res_acc} and \ref{fig_attn_conv_acc} demonstrate the training/validation accuracy and loss of models. Figure~\ref{fig_vit_res_dist} and \ref{fig_attn_conv_dist} demonstrate the Wasserstein distance of the persistence diagrams between adjacent layers of models.

\end{document}